\documentclass[
]{ceurart}

\usepackage{listings}
\usepackage{bbm}
\usepackage{ifpdf}

\begin{document}

\copyrightyear{2025}
\copyrightclause{Copyright for this paper by its authors.
  Use permitted under Creative Commons License Attribution 4.0
  International (CC BY 4.0).}

\conference{CLEF 2025 Working Notes, 9 -- 12 September 2025, Madrid, Spain}

\title{BIBERT-Pipe on Biomedical Nested Named Entity Linking at BioASQ 2025}

\title[mode=sub]{Notebook for the MSM Lab at CLEF 2025}

\tnotemark[1]
\tnotetext[1]{You can use this document as the template for preparing your
  publication. We recommend using the latest version of the ceurart style.}

\author[1]{Chunyu Li}[%
orcid=0009-0000-3036-0275,
email=li.chunyu0412@gmail.com,
url=https://eeyorelee.github.io/,
]
\cormark[1]
\fnmark[1]

\author[1]{Xindi Zheng}[%
orcid=0009-0007-0974-788X,
email=xindizhe@gmail.com,
url=https://github.com/momoxia,
]
\fnmark[1]

\author[1]{Siqi Liu}[%
orcid=0009-0001-0372-7155,
email=liusiqisq0412@163.com,
url=https://github.com/cmwls,
]
\fnmark[1]

\address[1]{Individual Researcher}

\cortext[1]{Corresponding author.}
\fntext[1]{These authors contributed equally.}

\begin{abstract}
Entity linking (EL) for biomedical text is typically benchmarked on \emph{English-only corpora with flat mentions}, leaving the more realistic scenario of \emph{nested} and \emph{multilingual} mentions largely unexplored.  
We present our system for the \textbf{BioNNE 2025 Multilingual Biomedical Nested Named Entity Linking} shared task (English \& Russian), closing this gap with a lightweight pipeline that keeps the original EL model intact and modifies only three task-aligned components:
\textbf{Two-stage retrieval-ranking.} We leverage the same base encoder model in both stages: the retrieval stage uses the original pre-trained model, while the ranking stage applies domain-specific fine-tuning.
\textbf{Boundary cues.} In the ranking stage, we wrap each mention with learnable \texttt{[Ms]} / \texttt{[Me]} tags, providing the encoder with an explicit, language-agnostic span before robustness to overlap and nesting.
\textbf{Dataset augmentation.} We also automatically expand the ranking training corpus with three complementary data sources, enhancing coverage without extra manual annotation.  
On the BioNNE 2025 leaderboard, our two stage system, bilingual bert (\textbf{BIBERT-Pipe}), ranks \textbf{third} in the multilingual track, demonstrating the effectiveness and competitiveness of these minimal yet principled modifications.  
Code are publicly available at \url{https://github.com/Kaggle-Competitions-Code/BioNNE-L}.

\end{abstract}


\maketitle

\section{Introduction}

Biomedical entity linking (BEL) -- also known as named entity normalization or grounding -- is the task of mapping entity mentions in the text to entries in a reference knowledge base. In the biomedical domain, EL plays a vital role in text mining by standardizing mentions of diseases, genes, drugs, and other entities to canonical identifiers \citep{french2023overview}. This normalization resolves synonymy and ambiguity: for example, the abbreviation “WSS” could refer to Wrinkly Skin Syndrome or Weaver-Smith Syndrome, and linking it to the correct concept ID disambiguates the intended meaning \citep{Garda2023}. By grounding mentions to KB concepts (e.g., UMLS or Wikidata entries), EL enables effective information integration, improves literature search (e.g., concept-based PubMed indexing), and facilitates downstream tasks such as relation extraction and question answering.

While early BEL research has made significant progress in English-only settings with flat (non-overlapping) mentions, real-world biomedical documents often exhibit \textit{nested entities} and appear in \textit{multiple languages}—posing persistent challenges that remain under-addressed.

Nested mentions—where one entity is embedded within or overlaps another—are prevalent in biomedical literature. For example, in \textit{“EGFR exon 19 deletion mutation”}, the terms \textit{“EGFR”} and \textit{“exon 19 deletion”} refer to distinct concepts, both requiring normalization. Ignoring nested structures can lead to incomplete or incorrect linking. Meanwhile, the increasing volume of biomedical text in non-English languages highlights the importance of multilingual BEL. Studies have shown that models trained in English exhibit significant performance drops when applied to languages like Spanish or Russian~\citep{guven2023multilingual, liu2021xl}.

Several technical barriers exacerbate these challenges: (i) the lack of annotated multilingual data, especially in low-resource biomedical languages; (ii) inconsistencies in concept coverage across languages in knowledge bases; and (iii) the inherent ambiguity and granularity of biomedical terminology. Existing EL pipelines are typically not equipped to handle these complexities simultaneously.

In this paper, we propose a lightweight, encoder-agnostic pipeline for multilingual, nested biomedical EL. Our method introduces three key enhancements: (i) a two-stage retrieval-ranking strategy that leverages the same base encoder model, where the retrieval stage utilizes the original pre-trained model and the ranking stage benefits from contrastive learning training; (ii) boundary cue tagging, using learnable tokens (\texttt{[Ms]} / \texttt{[Me]}) to explicitly encode span boundaries, enabling robust modeling of nested mentions; and (iii) dataset augmentation by incorporating additional complementary data sources, enriching training coverage without requiring manual annotation. Our approach maintains the original EL model architecture while significantly improving robustness across languages and nested spans. It can be seamlessly integrated with several biomedical encoders (e.g., BioLinkBERT~\cite{yasunaga2022linkbert}, SapBERT~\cite{liu2021sapbert}) and adapted to multilingual scenarios with minimal overhead.

Our system achieved third place in the BioNNE-L 2025 multilingual track~\cite{BioASQ2025BioNNEl}, demonstrating that our proposed techniques—two-stage retrieval ranking, boundary cue tagging, and data augmentation—are not only lightweight and effective but also highly generalizable. They can be seamlessly applied to a variety of base encoders and readily integrated into multilingual biomedical EL systems. This highlights the practical value of our approach for building robust, scalable solutions to cross-lingual entity linking tasks.

\begin{figure}[b]
  \centering
  \includegraphics[width=0.571\linewidth]{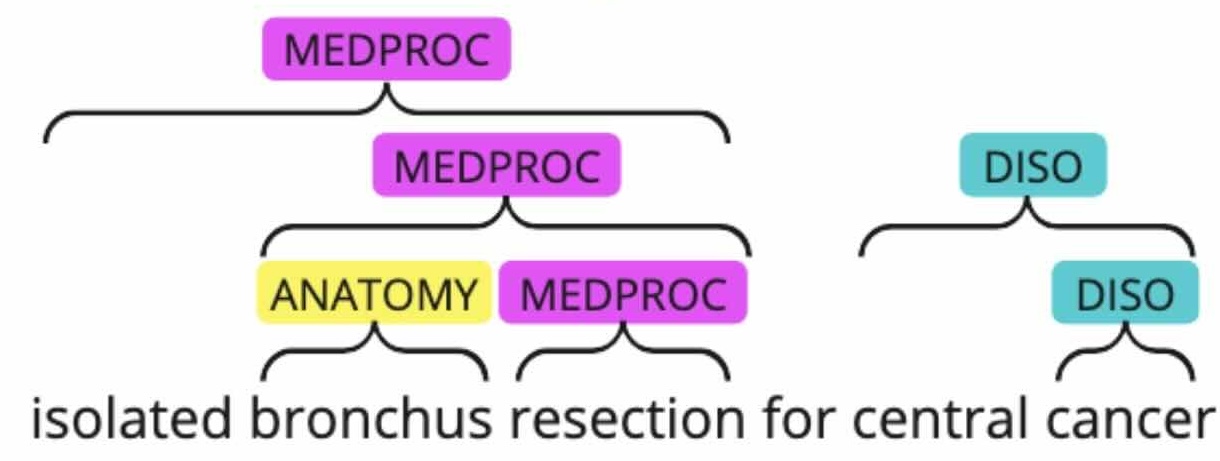}
  \caption{\textbf{ Example of nested named entities in NEREL-BIO.}
  The English phrase \emph{“isolated bronchus resection for central cancer”}
  is annotated with overlapping spans:
  the outer span (magenta) is a diagnostic procedure (\textsc{medproc});
  inside it, the token \emph{bronchus} is an  anatomical structure (yellow,
  \textsc{anatomy}), while \emph{resection} is again a procedure
  (magenta, \textsc{medproc}).
  A separate right-hand branch shows the phrase \emph{central cancer},
  where both the full span and the nested core \emph{cancer} are labelled
  as a disease (cyan, \textsc{diso}).
  This illustrates the two challenges of the task: \emph{nesting}
  (entities contained within entities) and
  \emph{fine-grained, type-specific} normalisation.}
  \label{fig:task_description}
\end{figure}

\section{Task Overview}

To further advance research in biomedical entity linking, BioASQ 2025~\cite{BioASQ2025overview} holds a task, BioNNE-L~\cite{BioASQ2025BioNNEl}: Nested NER in Russian and English. The BioNNE-L shared task focuses on NLP challenges in entity linking, also known as medical concept normalization (MCN), for English and Russian languages. The goal is to map biomedical entity mentions to a comprehensive set of medical concept names and their concept unique identifiers (\textsc{Cui}s) from the UMLS.  The train, dev, and test datasets include mentions of disorders, anatomical structures, and chemicals, all mapped to concepts from the UMLS. The BioNNE-L task utilizes the MCN annotation of the NEREL-BIO dataset~\cite{NERELBIO}, which provides annotated mentions of disorders, anatomical structures, chemicals, diagnostic procedures, and biological functions.

\section{Related Work}
\paragraph{Multilingual Biomedical Entity Linking.}
Multilingual BEL is an increasingly important research direction due to the global nature of biomedical literature. Traditional approaches often rely on translation to English prior to linking, but this can introduce noise and domain mismatch~\citep{liu2021xl}. To overcome these limitations, recent work has focused on cross-lingual encoders and alignment techniques. SapBERT~\citep{liu2021sapbert} uses self-alignment pretraining with UMLS synonym pairs across languages to learn language-agnostic biomedical embeddings. Guven and Lamurias~\citep{guven2023multilingual} study bi-encoder models on English and Spanish corpora and highlight persistent performance gaps on non-English datasets. 

\paragraph{Nested Mention Normalization.}
Nested named entities are a known challenge for EL systems. Standard EL models often assume flat mention boundaries and can not resolve overlapping entities. The MCN dataset~\citep{NERELBIO} extends entity linking to nested mentions in both English and Russian, providing a valuable benchmark. However, few EL systems explicitly model nested mentions. Some recent work, such as Con2GEN~\citep{zhu2023controllable} addresses multilingual biomedical entity linking using a generation-based approach with predefined prompts, effectively capturing dependencies between mentions and concepts. However, such generative methods may involve increased model complexity and computational resources compared to discriminative approaches.

\paragraph{Contrastive and Graph-Based Learning.}
Contrastive learning has proven effective for biomedical EL, particularly in bi-encoder architectures. GEBERT~\cite{sakhovskiy2023graph} combines a Transformer with a graph neural encoder over the UMLS knowledge graph. 
It aligns graph node embeddings with textual descriptions through node-text contrastive learning. BERGAMOT~\cite{sakhovskiy2024biomedical} extends this with multiple contrastive losses and multilingual pretraining, improving generalization across languages and domains. SapBERT~\citep{liu2021sapbert} trains with an InfoNCE loss to align mention and concept representations. BERGAMOT~\citep{sakhovskiy2024biomedical} extends this with multilingual graph-based contrastive learning, incorporating ontology structure. Con2GEN~\citep{zhu2023controllable} instead adopts a controllable generation strategy to bridge mention-concept alignment using cross-lingual templates. While these methods demonstrate strong results, they often require complex training setups or extensive graph preprocessing.

\paragraph{Large Language Models.} 
LLMs like ChatGPT and GPT-4 have been tested on biomedical entity link tasks~\cite{jahan-etal-2023-evaluation}. 
While flexible, they often underperform domain-specific fine-tuned models in complex scenarios ~\cite{chen-etal-2025-benchmarking}. 
Instruction tuning and prompt engineering have been explored to close this gap~\cite{ding-etal-2024-chatel}, but performance is still limited without task-specific adaptation.

\paragraph{Our Contribution.}
In contrast to prior work, our method is efficient and explicitly designed for both multilinguality and nesting. It requires no architectural change to the encoder and is compatible with any transformer-based biomedical model. Our use of span boundary cues provides strong supervision for nested and cross-lingual linking, while dataset augmentation further improves accuracy.

\section{Method}
\label{sec:method}

Our approach follows a two-stage paradigm:
(\emph{i}) dense retrieval to obtain a small set of plausible concepts for each
mention, and  
(\emph{ii}) cross–encoder ranking to pick the best concept.
Although the backbone encoder may vary, the surrounding pipeline
remains unchanged and is illustrated in Figure~\ref{fig:rank}.

\subsection{Formal Definition}
\label{sec:formal}

Let $\mathcal{K}=\{\,c_{1},\dots,c_{|\mathcal{K}|}\,\}$ be a biomedical
knowledge base whose entries are represented by canonical names
and concept unique identifiers (\textsc{Cui}s).  
Given a document $D$ written in language
$\ell\!\in\!\{\textsc{en},\textsc{ru}\}$ that contains a set of (possibly
\emph{nested}) entities mentioned in
$\mathcal{M}(D)=\{\,m_{1},\dots,m_{N}\,\}$,
The goal is to find a mapping
\[
\Phi : (m_{i},D,\ell) \longrightarrow c^{\star}\in\mathcal{K},
\qquad 1\le i\le N,
\]
where $c^{\star}$ denotes the concept that
is \emph{semantically equivalent} to the surface form of $m_{i}$ in its
context.

We factor $\Phi$ into two components:
\vspace{-0.25\baselineskip}
\begin{align}
\textbf{Retrieval:}\quad &
  f_{\text{ret}}(m_i,D,\ell)\;\longrightarrow\;
  C_i = \bigl\langle c_i^{1},\dots,c_i^{k}\bigr\rangle,
  \label{eq:retrieval}\\[2pt]
\textbf{Rank:}\quad &
  f_{\text{rank}}\!\bigl(m_i,D,C_i,\ell\bigr)\;\longrightarrow\;
  \hat{c}_i \in C_i ,
  \label{eq:rank}
\end{align}
where \(k\!\ll\!|\mathcal{K}|\) (we use \(k=10\)).
Let \(c_i^{\star}\) be the gold concept for mention \(m_i\).
We report:

\[
\text{Acc@1}= \frac{1}{N}\sum_{i=1}^{N}\mathbbm{1}\bigl[\hat{c}_i = c_i^{\star}\bigr],\qquad
\text{Acc@}k = \frac{1}{N}\sum_{i=1}^{N}\mathbbm{1}\bigl[c_i^{\star}\in \text{Top-}k(\hat{\pi}_i)\bigr],\;k\in\{5,10\},
\]

where \(\hat{\pi}_i\) is the ranker–sorted list of the
\(k\) retrieval candidates for the \(i\)-th mention.
Thus, the rank stage simply re-orders the 10 candidates returned by the retrieval stage, and
we evaluate whether the gold concept appears within the first \(k\) positions.

\subsection{Framework}
\label{sec:framework}

\paragraph{Retrieval stage.}
We experiment with five publicly available biomedical encoders, including \emph{BioLink\-BERT}~\cite{yasunaga2022linkbert} and \emph{Bio\-med\-BERT}~\cite{pubmedbert}(  \texttt{abstract/fulltext}, \emph{bert-base-uncased}).
For every mention, we wrap its span with our boundary cues
\texttt{[Ms]} and \texttt{[Me]},
encode the sequence, and compute a cosine similarity to
all concept representations in $\mathcal{K}$.\footnote{Concept vectors are
pre-computed once per encoder.}

\begin{figure}[htbp]              
  \centering
  \includegraphics[width=\linewidth]{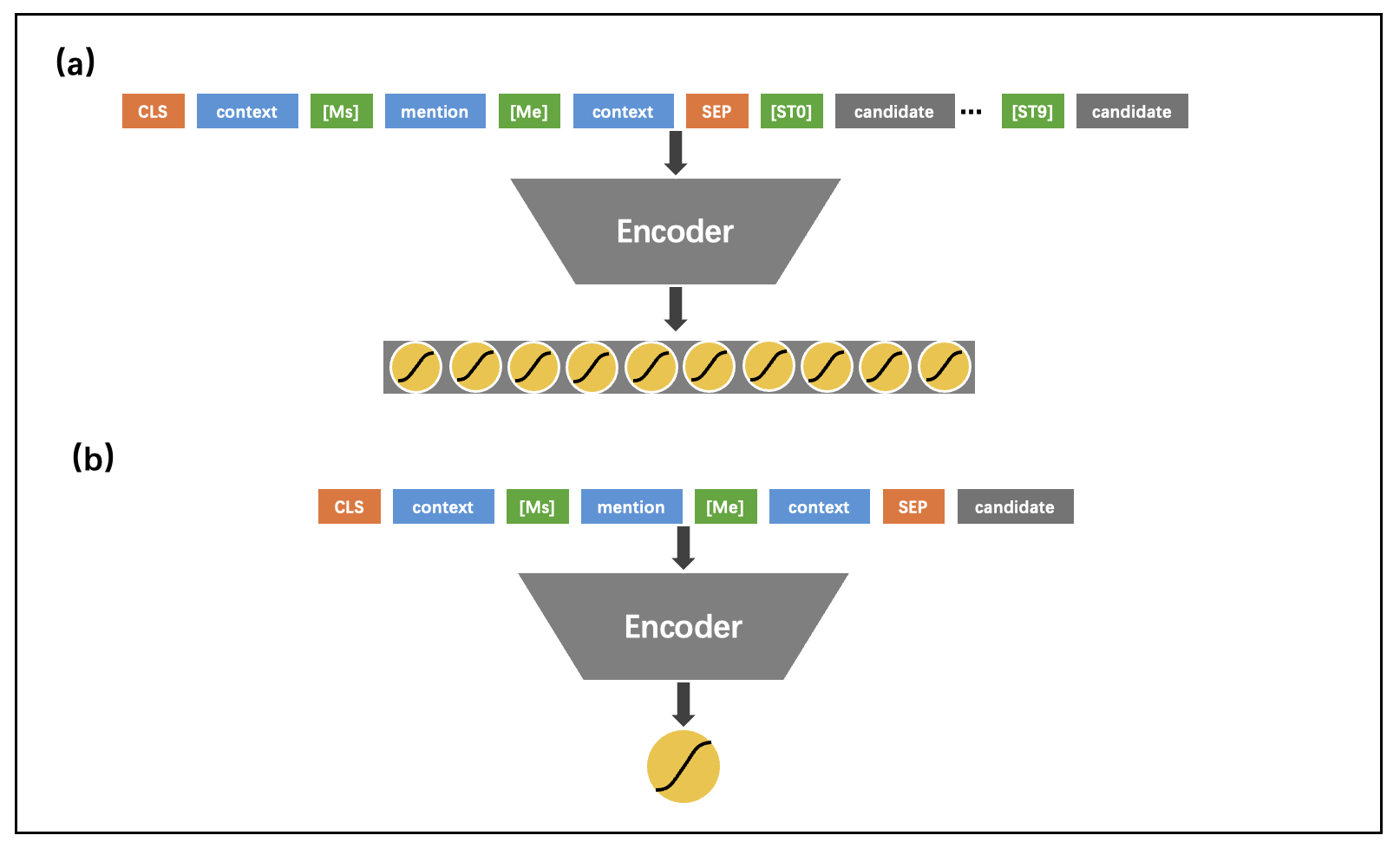}
    \caption{
    \textbf{Pipeline of our rank stage.}
    \textbf{(a) Listwise} – the $k$ candidates returned by the retrieval stage are \emph{concatenated} after the \texttt{[SEP]} token (\texttt{[ST0]}\dots\texttt{[ST${}_{k-1}$]}\,), so a \emph{single} forward pass of the encoder produces $k$ logits, one for each candidate.  
    \textbf{(b) Contrastive Learning} – each candidate is paired with the mention context in an independent input sequence; the encoder is applied $k$ times and outputs a binary logit for every candidate individually.
    }
  \label{fig:rank} 
\end{figure}

\paragraph{Rank stage.}
After the retrieval stage, we need to rank the \textsc{Cui}s. In particular, we build our ranking model without extra model modification. It is efficient to train the rank model with the retrieval model. We use two types of architecture to feed k candidates from the retrieval stage:

\begin{enumerate}
\item \emph{Listwise~(LTR)} (Figure~\ref{fig:rank}\,a):
the $k$ candidates $\langle c_{i}^{1}\!\dots c_{i}^{k}\rangle$
are concatenated after the \texttt{[SEP]} token;
One forward pass yields $k$ logits
$\mathbf{z}\!\in\!\mathbb{R}^{k}$ trained with a listwise
soft margin loss.

\item \emph{Contrastive Learning~(CL)} (Figure~\ref{fig:rank}\,b):
Each candidate is paired with the mention context
and processed independently;
every pass produces a binary score
$z\!\in\!\mathbb{R}$ optimized by cross-entropy.
\end{enumerate}

Since the LTR scheme processes all $k$ candidates in a single forward pass, while the CL scheme handles only one candidate per pass, the CL scheme inherently requires $k$ times more computation during both training and inference. Despite costing $k$ times more computation, the CL scheme eliminates cross-candidate interference. Obviously, each forward pass evaluates a single candidate against the mention context, reducing the problem to an independent binary decision that the model can learn more easily.

\subsection{Data Augmentation.}
Considering the limit of the training set, we also add additional dataset for the RU and BI tracks, including \textbf{MedMentions}~\cite{mohan2019medmentionslargebiomedicalcorpus}, a manually annotated resource
for the recognition of English biomedical concepts, and \textbf{MCN}~\cite{NERELBIO}, a novel dataset for nested entity linking in Russian. We reformat these two datasets to suit the competition and only keep the three entity types: (i) Disease (DISO), (ii) Chemical (CHEM), (iii) Anatomy (ANATOMY).

\section{Experiment}
\paragraph{Datasets.}  
We follow the official \texttt{BioNNE 2025} split: the train set with extra dataset is for training the model, the \textit{development set} is used for model selection; the \textit{evaluation set} is kept blind for final ranking.  
We set the retrieval numbers to \(k=10\). The base dataset of this task is NEREL-BIO ~\cite{NEREL-BIO-COLING-2024}.

\paragraph{Metrics.}  
For retrieval we report \(\text{Acc@}k\);  for ranking we use the cross-validated \(\text{Acc@}1\) (\(\text{CV Acc}\)) on the development folds.  
Final leaderboard numbers are the organisers’ \(\text{Acc@1}\) on the hidden evaluation set.

\subsection{Retrieve Stage}

\begin{table}[!ht]
  \caption{Retrieval accuracy (\textbf{Acc@k}) of base encoders on the development sets for English (EN), Russian (RU), and Bilingual (BI).}
  \label{tab:retrieval-combined}
  \centering
  \footnotesize
  \setlength{\tabcolsep}{4pt}
  \begin{tabular}{lccc}
    \toprule
    Model & Acc@1 & Acc@5 & Acc@10 \\
    \midrule
    \multicolumn{4}{l}{\textbf{English (EN) dev set}} \\
    \midrule
    GEBERT              & 0.5898 & 0.7654 & 0.7979 \\
    BioLinkBERT-large             & 0.4270 & 0.6030 & 0.6210 \\
    BioLinkBERT-base              & 0.4720 & 0.6530 & 0.6710 \\
    SapBERT--PubMedBERT (fulltext) & \textbf{0.6115} & 0.7698 & 0.8043 \\
    SapBERT--PubMedBERT (mean-token) & 0.6038 & \textbf{0.7723} & \textbf{0.8184} \\
    \midrule
    \multicolumn{4}{l}{\textbf{Russian (RU) dev set}} \\
    \midrule
    SapBERT-UMLS-XLMR (base)  & 0.4914 & 0.5497 & 0.5686 \\
    SapBERT-UMLS-XLMR (large) & \textbf{0.5103} & \textbf{0.5613} & \textbf{0.5763} \\
    \midrule
    \multicolumn{4}{l}{\textbf{Bilingual (BI) dev set}} \\
    \midrule
    SapBERT-UMLS-XLMR (base)  & 0.5197    & 0.7021    & 0.7331    \\
    SapBERT-UMLS-XLMR (large) & \textbf{0.5389} &\textbf{ 0.7171} & \textbf{0.7500} \\
    \bottomrule
  \end{tabular}
\end{table}

Table~\ref{tab:retrieval-combined} compares six off-the-shelf
biomedical encoders in three tracks.  
On the English track, \textbf{SapBERT--PubMedBERT (fulltext)} gives the
strongest \(\text{Acc@1}=0.6115\), while its
\textit{mean-token} pooling variant slightly improves the recall
(\(\text{Acc@10}=0.8184\)).  
Russian retrieval is notably harder: even
\textbf{SapBERT–XLMR–large} reaches only \(0.5103\) Acc@1.  
The bilingual runs averages those behaviours, yielding
\(0.5389\) Acc@1 at best.  
These numbers indicate that the retrieval module already places the gold
concept within the top-10 for more than 80\%(EN) and more than 75\%(BI)
of mentions, leaving ample head-room for a ranker.

\subsection{Rank Stage}
For the ranking phase we keep the best retriever for each track:
\textbf{SapBERT–PubMedBERT (mean-token)} for English, and
\textbf{SapBERT–UMLS–XLMR (large)} for both the Russian and Bilingual tracks.

Table~\ref{tab:rank-english} analyses two cross-encoder architectures on the English dev set.  
The \textbf{Listwise}, although computationally cheap (one
forward pass), plateaus at \(0.5918\) CV Acc.  
Switching to the \textbf{CL} design—i.e.\ an independent binary
decision per candidate—raises accuracy to
\(0.6604\) when additional \textit{MedMentions} are used.
We attribute the gain to two factors: (i) candidates no longer compete inside the softmax, thus reducing interference, and (ii) the binary objective is simpler, allowing the model to specialise on fine-grained lexical cues.

Bilingual results in Table~\ref{tab:rank-results} confirm the trend.  Incorporating \textit{MedMentions} and the MCN dataset adds a further \(0.7\)–\(2.4\) pp on Russian and bilingual tracks, but English still
benefits the most (\(+6.9\) pp).

\begin{table*}[!ht]
  \caption{
    Rank results on the English (EN) track.  
    The candidate set produced by the retrieval stage is identical for all systems
    (\textbf{Acc@1 = 0.6115}, \textbf{Acc@5 = 0.7698}, \textbf{Acc@10 = 0.8043});
    we therefore report only the cross-validated accuracy (\textbf{CV Acc}) of each ranker.
    “Listwise” vs.\ “CL” denotes two different architectures. For all of the experiments in this table, the post training epoch for ranking is 5, and the learning rate is chosen from 7e-6 or 1e-5.
  }
  \label{tab:rank-english}
  \centering\footnotesize
  \setlength{\tabcolsep}{5pt}
  \begin{tabular}{lccc}
    \toprule
    \textbf{Base Model} &
    \textbf{CV Acc} &
    \textbf{Approach} &
    \textbf{Training Data / $k$} \\
    \midrule
    BioLinkBERT-base          & 0.5871 & LTR& train, $k{=}5$ \\
    BioLinkBERT-large         & 0.5699 & LTR& train, $k{=}5$ \\
    BERT-base-uncased         & 0.5683 & LTR& train, $k{=}5$ \\
    BiomedBERT-abstract       & 0.5871 & LTR& train, $k{=}5$ \\
    BiomedBERT-abstract       & 0.5918 & CL    & train, $k{=}5$ \\
    BiomedBERT-abstract       & \textbf{0.6604} & CL & MedMentions + train, $k{=}5$ \\
    KRISSBERT                 & 0.6576 & CL    & MedMentions + train, $k{=}5$ \\
    BiomedBERT-fulltext       & 0.6536 & CL    & MedMentions + train, $k{=}5$ \\
    BiomedBERT-fulltext       & 0.6532 & CL    & MedMentions + train, $k{=}10$ \\
    \bottomrule
  \end{tabular}
\end{table*}

\begin{table*}[!ht]
\caption{Rank stage results on English (EN), Russian (RU), and Bilingual (BI) tracks. Details include the training dataset for the rank stage. For all of experiments of this table, the further training epoch for ranking is 5 and the learning rate is chosen from 7e-6 or 1e-5.}
\label{tab:rank-results}
\centering\small
\setlength{\tabcolsep}{4pt}
\begin{tabular}{lcccccl}
\toprule
\textbf{Base Model} & \textbf{Lang} & \textbf{CV(Acc)}  & \textbf{Approach} & \textbf{Details} \\
\midrule
BiomedNLP-BiomedBERT-base-uncased-abstract                & EN & 0.5918      & CL           & train\\
BiomedNLP-BiomedBERT-base-uncased-abstract                & EN & 0.6604      & CL      & train+MedMentions\\
\midrule
SapBERT-XLMR-large                & RU & 0.6131 & CL           & train \\
SapBERT-XLMR-large                & RU & 0.6204 & CL           & train+MedMentions+MCN,\\

\midrule
SapBERT-XLMR-large                & BI & 0.6083 & CL           & train\\
SapBERT-XLMR-large                & BI & 0.6319 & CL           & train+MedMentions+MCN\\

\bottomrule
\end{tabular}
\end{table*}

\subsection{Final result}
Table~\ref{tab:final-results} summarises leaderboard scores.
Our best submissions  \textbf{SapBERT-XLMR-large + MedMentions + MCN} for RU/BI, and \textbf{BiomedBERT-abstract + MedMentions} for EN achieve \(\mathbf{0.6497}\), \(\mathbf{0.6370}\) and
\(\mathbf{0.6370}\) Acc@1 on RU, BI and EN, respectively, ranking third overall in the bilingual track.

\begin{table*}[!ht]
\caption{Final results on English (EN), Russian (RU), and Bilingual (BI) tracks on evaluation dataset.  For all of experiment of this table, the further training learning rate is chosen from 7e-6 or 1e-5.}
\label{tab:final-results}
\centering\small
\setlength{\tabcolsep}{4pt}
\begin{tabular}{lcccccl}
\toprule
\textbf{Base Model} & \textbf{Lang} & \textbf{Acc} & \textbf{Approach} & \textbf{Details} \\
\midrule
BiomedBERT-abstract               & EN & 0.6197 & CL & MedMentions+train, epoch=2 \\
BiomedBERT-abstract               & EN & 0.6273 & CL & MedMentions+train+dev,  epoch=2 \\
SapBERT-XLMR-large                & EN & 0.6370 & CL & train+MedMentions+dev,  epoch=1   \\
\midrule
SapBERT-XLMR-large                & RU & 0.6452 & CL & train, epoch=5\\
SapBERT-XLMR-large                & RU & 0.6497 & CL & train+MedMentions+MCN+dev, epoch=5\\
\midrule
SapBERT-XLMR-large                & BI & 0.6229 & CL & train, epoch=1 \\
SapBERT-XLMR-large                & BI & 0.6342 & CL & train+MedMentions+MCN+dev, epoch=1\\
\bottomrule
\end{tabular}
\end{table*}

\section{Ablation Study}
\subsection{Boundary Cues}
To assess the contribution of boundary cues, we also perform an ablation study Table~\ref{tab:special-tokens} on boundary cues, known as special tokens [Ms] and [Me], which indicate the start and end of the target entity. 
The improvement is most pronounced on the Russian (RU) track, where the Acc@1 increases by \(6.60\%\).  We attribute this to the richer morphology of Russian: the explicit \texttt{[Ms]} / \texttt{[Me]} markers help the model to delineate entity spans that may otherwise be obscured by inflectional endings.  For the English (EN) and the Bilingual (BI) setting, the gains are more modest \(1.20\%\) and \(1.24\%\), respectively - but still positive, confirming that boundary information remains beneficial even in languages with a relatively simpler morphology.

\begin{table*}[!ht]
  \centering\footnotesize
  \setlength{\tabcolsep}{5pt}
  \caption{Ablation study on the effect of boundary cues for the English (EN), Russian (RU), and Bilingual (BI) tracks of the evaluation set. All experiments are conduct with the same hyper-parameters on the SapBERT-XLMR-large base model; the only difference is whether the boundary-cue tokens are included. Performance is reported using Acc@1, consistent with our earlier experiments.}
  \label{tab:special-tokens}
  \begin{tabular}{lccc}
    \toprule
\textbf{Lang} &
\textbf{w/ [Ms] and [Me]} &
\textbf{w/o [Ms] and [Me]} &
\textbf{Gain}\\
\midrule
EN & 0.6370 & 0.6292 & $0.0078$ ($1.24\%$) \\
RU & 0.6497 & 0.6095 & $0.0402$ ($6.60\%$) \\
BI & 0.6342 & 0.6267 & $0.0075$ ($1.20\%$) \\
\bottomrule
\end{tabular}
\end{table*}

\section{Conclusion}
\label{sec:conclusion}

We present a simple yet effective two–stage pipeline for the
\textsc{BioNNE 2025} Bilingual Nested Entity Linking task.
Keeping the \emph{base encoder untouched}, we obtained
competitive performance by addressing three task-specific
bottlenecks: (i)~explicit mention boundary cues
(\texttt{[Ms]/[Me]}) indicating the position of mention, (ii)~efficient rank architecture design for ranking mention and
(iii)~data augmentation with MedMentions / MCN boosting the final result.
On the official leaderboard our system ranks \textbf{3\textsuperscript{rd}}
in BI track, with \text{Acc@1} of \(0.637\) (BI),
while training on a single Nvidia 3090.

\section{Declaration on Generative AI}
During the preparation of this work, the authors used ChatGPT and Grammarly, to:
Grammar and spelling check, paraphrase, and minor translation.
After using these tools, the authors reviewed and edited the content as needed and assume full responsibility for the content of the publication.

%%
%% Define the bibliography file to be used
\bibliography{bibert-pipe}

\appendix

\end{document}